\documentclass[11pt,a4paper]{article}
\usepackage{authblk}
\usepackage{blindtext}
\usepackage[hyperref]{acl2018}
\usepackage{times}
\usepackage{latexsym}
\usepackage{enumitem}
\usepackage{graphicx}
\usepackage{algorithm}
\usepackage{amsmath}
\usepackage{mathtools}
\usepackage{algorithmicx}

\usepackage{url}
\usepackage{hyperref}
\aclfinalcopy % Uncomment this line for the final submission
\title{Natural Language Interaction with Explainable AI Models}

\author[1]{Arjun R Akula}
\author[2]{Sinisa Todorovic}
\author[3]{Joyce Y Chai}
\author[4]{Song-Chun Zhu}
\affil[1,4]{University of California, Los Angeles}
\affil[2]{Oregon State University}
\affil[3]{Michigan State University}
\affil[1]{\tt aakula@ucla.edu}
\affil[2]{\tt sinisa@oregonstate.edu}
\affil[3]{\tt jchai@cse.msu.edu}
\affil[4]{\tt sczhu@stat.ucla.edu}

\date{}
\begin{document}

\maketitle
\begin{abstract}
 This paper presents an explainable AI (XAI) system that provides explanations for its predictions. The system consists of two key components -- namely, the prediction And-Or graph (AOG) model for recognizing and localizing concepts of interest in input data, and the XAI model for providing explanations to the user about the AOG's predictions. In this work, we focus on the XAI model specified to interact with the user in natural language, whereas the AOG's predictions are considered given and represented by the corresponding parse graphs (pg's) of the AOG. Our XAI model takes pg's as input and provides answers to the user's questions using the following types of reasoning: direct evidence (e.g., detection scores), part-based inference (e.g., detected parts provide evidence for the concept asked), and other evidences from spatiotemporal context (e.g., constraints from the spatiotemporal surround). We identify several correlations between user's questions and the XAI answers using Youtube Action dataset.
\end{abstract}

\section{Introduction}
An explainable AI (XAI) model aims to provide transparency (in the form of justification, explanation, etc) for its predictions or actions made by it~\cite{baehrens2010explain,lipton2016mythos,ribeiro2016should,miller2017explanation,biran17explanation,biran2017human}. Recently, there has been a lot of focus on building XAI models, especially to provide explanations for understanding and interpreting the predictions made by deep learning models (e.g. explaining models in medical diagnosis domains~\cite{hatamizadeh2018automatic, hatamizadeh2019deep}).

\begin{figure}
\centering
  \includegraphics[width=0.9\columnwidth]{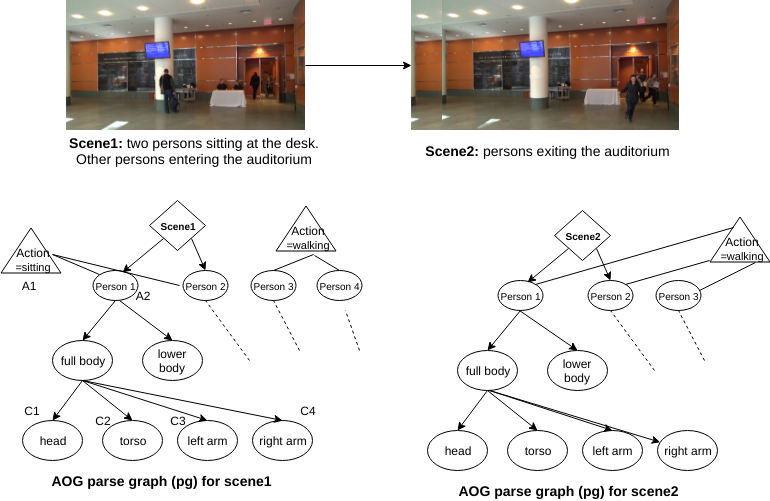}
  \caption{Two frames (scenes) of a video: (a) top-left image (scene1) shows two persons sitting at the reception and others entering the auditorium and (b) top-right (scene2) image people running out of an auditorium. Bottom-left shows the AOG parse graph (pg) for the top-left image and Bottom-right shows the pg for the top-right image}~\label{fig:figure1}
\end{figure}

Consider for example, two frames (scenes) of a video shown in Figure~\ref{fig:figure1}. An action detection model might predict that two people in the scene1 are in sitting posture. User might be interested to know more details about the prediction such as: Why do the model think the people are in sitting posture? Why not standing instead of sitting? Why two persons are sitting instead of one? The XAI models aim to generate explanations to these questions from different perspectives such as follows: ``\textit{action detection score for them to sit is higher than other actions such as standing}'', ``\textit{the torso, left\_arm and right\_arm poses of both the people suggest that they are in sitting pose}'', ``\textit{I found chairs behind the table in the beginning of the video and couldn't see them now, which is why I think they might be sitting on those chairs}''.

Explanations are considered to be interactive conversations~\cite{miller2017explanation,akula2013novel}. Therefore it is necessary to understand the underlying characteristics of such conversations. In this work, we propose a generic framework to interact with an XAI model in natural language. The framework consists of two key components -- namely, the prediction And-Or graph (AOG) model~\cite{zhu2007stochastic} for recognizing and localizing concepts of interest in input data, and the XAI model for providing explanations to the user about the AOG's predictions. 

The And-Or graph is a hierarchical and compositional representation recursively defined to capture contextual information. AOG structure embodies the expressiveness of context sensitive grammars and probabilistic reasoning of graphical models. Spatial, Temporal and Causal decomposition of entities and the relations between them can be modeled using AOG. In this work, we focus on the XAI model specified to interact with the user in natural language, whereas the AOG's predictions are considered given and represented by the corresponding parse graphs (pg's) of the AOG. Our XAI model takes pg's as input and provides answers to the user's questions using the following types of reasoning: direct evidence (e.g., detection scores), part-based inference (e.g., detected parts provide evidence for the concept asked), and other evidences from spatiotemporal context (e.g., constraints from the spatiotemporal surround). We created a new explanation dataset by using Youtube Action Videos dataset~\cite{liu2009recognizing}. To the best of our knowledge, this is the first explanation dataset that has explicit question and explanation pairs. We present several correlations between user's questions and the XAI answers using our explanation dataset.

%Firstly, we categorize different types of questions a user might ask about a model's prediction. We then categorize different ways in which an XAI model could generate explanations to answer user's questions. We believe that identifying the type of the input question and the type of the output explanation that needs to be generated is the most important step in providing an acceptable and convincing explanation to the user.

%Secondly, we present an And-Or graph (AOG) representation [] to capture the underlying evidences used by the XAI model in making a prediction. We show that the AOG representation facilitates the XAI model in generating context based explanations. Furthermore, we present a rule-based algorithm to predict the most appropriate explanation type for a given user question. We evaluate our approach on two explanation datasets: Visual Question Answering Explanation dataset (VQA-X) and Action Explanation dataset (ACT-X) []. We show that the proposed question and explanation categories are sufficient enough to represent all the variations of questions and explanations in the datasets. We also demonstrate that AOG representation is the key in generating most appropriate explanation for a given user question.  

\section{Related Work}
Several works~\cite{miller2017explanation} have been proposed in the past to understand the underlying characteristics of explanations. Lombrozo et al.~\cite{lombrozo2006structure} proposed that the explanations are typically contrastive: they account for one state of affairs in contrast to another. However the definitions of most of these explanation types are based on theoretical grounds~\cite{dennett1989intentional,chin2010background} and cannot be applied directly in practice. In our work, we propose explanation types that are motivated from an algorithmic perspective rather than on theoretical grounds. Hilton et al.~\cite{hilton1990conversational} proposed different types of contrastive questions that can be posed by the user to an XAI model such as  i) why X rather than not X; ii) why X rather than the default value for X; iii) why X rather than Y. In our experiments, we found a similar and more finer categorization to be helpful for analyzing the users questions. 

\section{XAI Question Types}
Questions posed by the user to obtain explanations from an XAI model are typically contrastive in nature~\cite{hilton1990conversational,lombrozo2006structure,miller2017explanation}. For example, questions such as ``Why does the model think the person is in sitting posture'', ``Why does the model think that two persons are sitting instead  of one?'', need contrastive explanations. In order to generate such explanations, we categorize the questions into the following 10 categories to understand the implicit contrast that a question presupposes:

\begin{enumerate}[leftmargin=*,noitemsep,topsep=0pt]

\item \textbf{WH-X} Contrastivity in these type of questions will be in the form of  ``Why X rather than not X''. For example, the question ``Why does the model think the person is sitting?'' is a  WH-X question based on the video shown in Figure~\ref{fig:figure1}. In this question, user wants to know the explanation as to why the person's action is predicted as sitting rather than not sitting. 
\item \textbf{WH-X-NOT-Y} Contrastivity in these type of questions will be in the form of  ``Why X rather than Y''. For example, the question ``Why does the model think the person is sitting and not standing?'' is a  WH-X-NOT-Y question. In this question, user wants to know the explanation as to why the person's action is predicted as sitting rather than standing. WH-X and WH-X-NOT-Y categories look similar and one might think they both need similar explanations. However, in our experiments, we found that they need different explanations.
\item \textbf{WH-X1-NOT-X2} Contrastivity in these type of questions will be in the form of  ``Why X1 rather than X2''. For example, the question ``Why does the model think two persons are sitting instead of three?'' is a  WH-X1-NOT-X2 question. It may be noted that WH-X-NOT-Y questions refer to the contrastivity between two different concepts X and Y whereas WH-X1-NOT-X2 refer to the contrastivity between two different observations about a single concept. 
\item \textbf{WH-NOT-Y} Contrastivity in these type of questions will be in the form of  ``Why not Y''. For example, the question ``Why does the model think the person is not standing?'' is a  WH-NOT-Y question. In this question, user wants to know the explanation as to why the person's action is not predicted as standing.
\item \textbf{NOT-X} User might want to correct the XAI model's understanding of a concept or argue with the XAI model over the validity of an evidence. For this purpose, we propose the question categories beginning with the prefix `NOT'. Questions of NOT-X category will be in the form of ``I think it is X rather than not X''. For example, the question ``I think the person is not sitting?'' is a  NOT-X question.
\item \textbf{NOT-X1-BUT-X2} Questions of NOT-X1-BUT-X2 category will be in the form of ``I think it is X1 rather than X2''. For example, the question ``I think there are two persons in the video and not just one'' is a  NOT-X1-BUT-X2 question.
\item \textbf{NOT-X-BUT-Y} Questions of NOT-X-BUT-Y category will be in the form of ``I think it is X rather than Y''. For example, the question ``I think the person is sitting and not standing'' is a  NOT-X-BUT-Y question.
\item \textbf{DO-X-NOT-Y} In some cases, user may want to see how system's interpretation/explanation would change by hiding/modifying certain evidences. For this purpose, we propose the question categories beginning with the prefix `DO'. Questions of DO-X-NOT-Y category will be in the form of  ``What if it is X rather than Y''. For example, the question ``What if the person is standing and not sitting?'' is a  DO-X-NOT-Y question. In this question, user wants to know how the system's interpretation of the scene in the video will change by assuming the person to be in the standing pose rather than in the sitting pose.
\item \textbf{DO-NOT-X} Questions of DO-NOT-X category will be in the form of  ``What if it is not X''. For example, the question ``What if the person is not sitting?'' is a  DO-NOT-X question.
\item \textbf{DO-X1-NOT-X2} Questions of DO-X1-NOT-X2 category will be in the form of  ``What if it is X1 and not X2''. For example, the question ``What if there are two persons in the video and not one?'' is a DO-X1-NOT-X2 question.
\end{enumerate}

\section{XAI Explanation Types}
Our XAI model takes AOG parse graphs (pg) as input and provides answers using the following six types of explanations. In our experiments, we found that these explanation types are sufficient to answer all the 10 different question types discussed in the previous section. 

\begin{enumerate}[leftmargin=*,noitemsep,topsep=0pt]
    \item \textbf{AOG Alpha explanation} Alpha explanation is the explanation generated by the XAI model using the direct evidence (e.g., detection scores). For example, consider the question ``Why does the model think that the person is sitting?''. Our XAI model, using the pg of scene1 shown in Figure~\ref{fig:figure1}, generates the following alpha explanation ``Action detection score for the person to sit is highest''. It may be noted that XAI model used the evidence from node A1 in the pg to generate this response without taking advantage of surrounding context (nodes).  
    \item \textbf{AOG Beta explanation} Beta explanation is the explanation generated by the XAI model using the part based inference (e.g., detected parts provide evidence for the concept asked). For example, consider the question ``Why does the model think that the object in the video is a person?''. Our XAI model, using the pg of scene1 shown in Figure~\ref{fig:figure1}, generates the following beta explanation ``Because I can see the person's head, torso, left arm and right arm''. It may be noted that XAI model used the evidence from nodes C1, C2, C3, C4 in the pg to explain the node A2. In other words, explanation for node A2 is generated by binding a subset of node A2's  child nodes in a bottom-up manner.
    \item \textbf{AOG Gamma explanation} Gamma explanation is the explanation generated by the XAI model using the top-down approach. For example, consider the question ``Why does the model think there is a chair in the video?''. Our XAI model, using the pg of scene1 shown in Figure~\ref{fig:figure1}, generates the following gamma explanation ``Because I can see a person in the sitting pose''. It may be noted that XAI model fails to generate alpha and beta explanations for this question as we cannot see any chair (node A3) in the video. However, using the evidence from the parent node A1 in the pg, XAI model is able to explain that there is a chair in the video.
    \item \textbf{Common-sense explanation} Sometimes, the evidence to generate an explanation may not be available in the pg. In these cases, XAI model needs to generate an explanation by contrasting the pg with common sense knowledge. In our experiments, we used an ontology to capture the common sense knowledge. 
    \item \textbf{Counterfactual explanation} XAI model may need to argue with the user over an evidence or explanation. For example, consider the question ``What if the person do not have head in the video?''. XAI model generates the following counterfactual response ``Is it possible for a person to exists without the head''. In our experiments, we found that these counterfactual responses play an important role to explain certain concepts to the user.  
    \item \textbf{Discourse based explanation} A discourse relation~\cite{carlson2003building,akula2019visual} tell us how two segments (or sentences) in the text/document are logically connected with each other. In particular, discourse relations provide the function of each text segment in the document. In this work, we extend the idea of the discourse phenomenon to videos (and images). The intuition behind this is simple: any video can be mapped to a document (or a group of sentences). And discourse can be used to explain coherence of any document. Therefore we can use discourse to explain coherence of a video. We believe that obtaining discourse cues from videos would help us in generating better context-sensitive explanations~\cite{akula2015novel}. 
    
    In our experiments, we used Rhetorical Structure Theory (RST)~\cite{carlson2003building} to capture discourse relations. In RST framework, the leaves correspond to text fragments called elementary discourse units (the minimal discourse units). The internal nodes of the tree correspond to contiguous text spans. Each node is characterized by its nuclearity, or essential unit of information and each node is also characterized by a rhetorical relation between two or more non-overlapping, adjacent text spans. For example, scene1 (where people are entering the auditorium) and scene2 (where people are exiting the auditorium) in the Figure~\ref{fig:figure1} are connected with the discourse relation `contrast', i.e. both these scenes compliment with each other. 
    
    In our XAI model discourse information plays an important role in generating context-rich explanations. For example, consider the user question ``I think people in scene1 are not entering the auditorium''. Our XAI model generates a counterfactual explanation ``That means people in scene2 shouldn't be coming out of auditorium''. Since scene1 and scene2 are connected with contrast discourse relation, our XAI model immediately inferred the inconsistency of the facts and generated the above counterfactual question.

\end{enumerate}

%make sure to add the point that if you already visited a parent node, its better to move to next best expln. you can take example of this from powerpoint slides.

\section{Experiments}
We created a new explanation dataset by using Youtube Action Videos dataset~\cite{liu2009recognizing}. These videos belong to 11 action categories such as basketball shooting, biking, diving, etc. For each action category, the videos are grouped into 25 groups with more than 4 action clips in it. For each domain, we manually collected a set of question and explanation pairs: one question for each question type, and five types of explanations for each question, i.e, a total of 550 question-explanation pairs. We annotated this dataset with help of 25 graduate students and 30 undergraduate students. We asked the students to choose the correct explanation for each question. Based on these annotations, we found several correlations among the question and explanation types.

Table~\ref{tab:table1} shows the percentage of explanation types chosen by the students for each of the question types. For the WH-X question type, we expected that most students would prefer to see the Alpha explanations. This is because, as WH-X question types are simple and straightforward, we thought direct evidence would be sufficient to explain them. However, we found that most students preferred Gamma explanations, i.e. explanations generated using top-down approach. This observation tells us that context based explanations are preferred by the humans even for the simpler questions. For WH-X-NOT-Y questions, some students preferred Alpha explanations and some preferred Gamma explanations. We found that dialogue/conversation history would be helpful in deciding between Alpha and Gamma explanations. 

\begin{table*}
  \centering
  \begin{tabular}{|l| r| r| r| r| r| r| r|}
    % \toprule
    \hline
    {\small\textit{Question type}}
    & {\small \textit{Alpha}}
      & {\small \textit{Beta}}
    & {\small \textit{Gamma}} 
    & {\small \textit{Counterfactual}} 
    & {\small \textit{Common sense}} 
    & {\small \textit{Discourse}} \\
    \hline
  
    WH-X & 23 & 3.9 & 46.2 & 19.2 & 3.8 & 3.8\\
WH-X1-NOT-X2 & 11.5 & 30.4 & 4.2 & 3.8 & 3.8 & 46.6\\
WH-X-NOT-Y & 36.5 & 2 & 34.6 & 3.8 & 15.4 & 7.6\\
WH-NOT-Y & 34.6 & 4.5 & 3.8 & 50.1 & 3.8 & 2.2\\
NOT-X & 26.9 & 0 & 7.7 & 42.3 & 19.2 & 0\\
NOT-X1-BUT-X2 & 26.9 & 0 & 0 & 53.8 & 15.4 & 0\\
NOT-X-BUT-Y & 3.8 & 26.9 & 3.8 & 65.4 & 0 & 0\\
DO-X-NOT-Y & 0 & 3.8 & 7.2 & 3.8 & 3.8 & 81.4\\
DO-NOT-X & 3.8 & 3.8 & 15.4 & 69.3 & 0 & 3.8\\
DO-X1-NOT-X2 & 3.8 & 3.8 & 8 & 65.2 & 7.2 & 8\\
\hline
    % \bottomrule
  \end{tabular}
  \caption{Percentage of explanation types (in columns) matched against the question types}~\label{tab:table1}
\end{table*}

For WH-X1-NOT-X2 questions, most of the students preferred Discourse based explanations and Beta explanations. This is expected because in these questions, users refer to multiple instances (i.e. X1, X2) of same entity (X). So either the discourse relations between these multiple instances or the part based relations of these instances would be helpful in generating the evidences. For WH-NOT-Y questions, majority of the students preferred common sense based reasoning. Again, this is consistent with our expectation because explaining why something is not observed needs more understanding of the domain. Discourse based explanation type is the most preferred explanation type for DO-X-NOT-Y question type. This is expected because discourse relations would be helpful in visualizing the bigger context of how evidences of different concepts can influence each other. Similar to WH-NOT-Y question type, common sense explanations are found to be the most preferred explanations for NOT-X, NOT-X1-BUT-X2, NOT-X-BUT-Y, DO-NOT-X and DO-X1-NOT-X2 question types.

\begin{figure}
\centering
  \includegraphics[width=0.9\columnwidth]{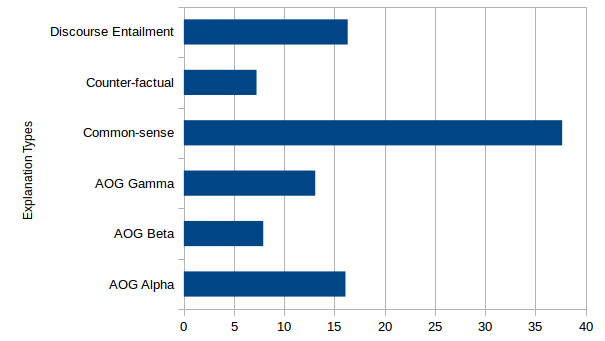}
  \caption{Summary of the total percentage of explanation types annotated across all the question types}~\label{fig:figure2}
\end{figure}

Figure~\ref{fig:figure2} summarizes the total number of explanation types that are annotated across all the question types. We may observe that Discourse based explanations and the Common sense explanations are the most sought-after explanations. This clearly illustrates the following two aspects: (a) necessity of capturing higher-level context and the coherence of the videos (or Images, documents, etc) through discourse relations, and (b) necessity for integrating richer domain knowledge in order to generate better common sense explanations. It may also be noted that AOG Alpha, Beta and Gamma explanations add up to more than 30\% of the explanations. That means AOG based prediction model is effective in capturing the underlying evidences of concepts.

\section{Conclusion}

In this paper, we proposed a generic framework to interact with Explainable AI (XAI) models in natural language. The proposed XAI model takes And-Or graph based parse graphs as input and provides answers to the user's questions using the following types of reasoning: direct evidence (e.g., detection scores), part-based inference (e.g., detected parts provide evidence for the concept asked), and other evidences from spatiotemporal context such as discourse relations, common sense reasoning, counterfactual reasoning, etc. In our experiments, we found that discourse based explanations and the common sense explanations are the most sought-after explanations. 
In the future, we plan to investigate the correlations between questions and explanation types by taking the dialogue history (question-explanation pairs from previous turns) into consideration.

% include your own bib file like this:
%\bibliographystyle{acl}
%\bibliography{acl2018}

\begin{thebibliography}{}
\expandafter\ifx\csname natexlab\endcsname\relax\def\natexlab#1{#1}\fi

\bibitem[{Akula et~al.(2013)Akula, Sangal, and Mamidi}]{akula2013novel}
Arjun Akula, Rajeev Sangal, and Radhika Mamidi. 2013.
\newblock A novel approach towards incorporating context processing
  capabilities in nlidb system.
\newblock In {\em Proceedings of the sixth international joint conference on
  natural language processing\/}. pages 1216--1222.

\bibitem[{Akula(2015)}]{akula2015novel}
Arjun~R Akula. 2015.
\newblock A novel approach towards building a generic, portable and contextual
  nlidb system.
\newblock {\em International Institute of Information Technology Hyderabad\/} .

\bibitem[{Akula and Zhu(2019)}]{akula2019visual}
Arjun~R Akula and Song-Chun Zhu. 2019.
\newblock Visual discourse parsing.
\newblock {\em arXiv preprint arXiv:1903.02252\/} .

\bibitem[{Baehrens et~al.(2010)Baehrens, Schroeter, Harmeling, Kawanabe,
  Hansen, and M{\~A}{\v{z}}ller}]{baehrens2010explain}
David Baehrens, Timon Schroeter, Stefan Harmeling, Motoaki Kawanabe, Katja
  Hansen, and Klaus-Robert M{\~A}{\v{z}}ller. 2010.
\newblock How to explain individual classification decisions.
\newblock {\em Journal of Machine Learning Research\/} 11(Jun):1803--1831.

\bibitem[{Biran and Cotton(2017)}]{biran17explanation}
Or~Biran and Courtenay Cotton. 2017.
\newblock Explanation and justification in machine learning: A survey.
\newblock In {\em IJCAI-17 Workshop on Explainable AI (XAI)\/}. page~8.

\bibitem[{Biran and McKeown(2017)}]{biran2017human}
Or~Biran and Kathleen McKeown. 2017.
\newblock Human-centric justification of machine learning predictions.
\newblock {\em IJCAI, Melbourne, Australia\/} .

\bibitem[{Carlson et~al.(2003)Carlson, Marcu, and
  Okurowski}]{carlson2003building}
Lynn Carlson, Daniel Marcu, and Mary~Ellen Okurowski. 2003.
\newblock Building a discourse-tagged corpus in the framework of rhetorical
  structure theory.
\newblock In {\em Current and new directions in discourse and dialogue\/},
  Springer, pages 85--112.

\bibitem[{Chin-Parker and Bradner(2010)}]{chin2010background}
Seth Chin-Parker and Alexandra Bradner. 2010.
\newblock Background shifts affect explanatory style: how a pragmatic theory of
  explanation accounts for background effects in the generation of
  explanations.
\newblock {\em Cognitive processing\/} 11(3):227--249.

\bibitem[{Dennett(1989)}]{dennett1989intentional}
Daniel~Clement Dennett. 1989.
\newblock {\em The intentional stance\/}.
\newblock MIT press.

\bibitem[{Hatamizadeh et~al.(2019)Hatamizadeh, Hosseini, Liu, Schwartz, and
  Terzopoulos}]{hatamizadeh2019deep}
Ali Hatamizadeh, Hamid Hosseini, Zhengyuan Liu, Steven~D Schwartz, and Demetri
  Terzopoulos. 2019.
\newblock Deep dilated convolutional nets for the automatic segmentation of
  retinal vessels.
\newblock {\em arXiv preprint arXiv:1905.12120\/} .

\bibitem[{Hilton(1990)}]{hilton1990conversational}
Denis~J Hilton. 1990.
\newblock Conversational processes and causal explanation.
\newblock {\em Psychological Bulletin\/} 107(1):65.

\bibitem[{Imran et~al.(2018)Imran, Hatamizadeh, Ananth, Ding, Terzopoulos, and
  Tajbakhsh}]{hatamizadeh2018automatic}
Abdullah Imran, Ali Hatamizadeh, Shilpa~P Ananth, Xiaowei Ding, Demetri
  Terzopoulos, and Nima Tajbakhsh. 2018.
\newblock Automatic segmentation of pulmonary lobes using a progressive dense
  {V}-network.
\newblock In {\em Deep Learning in Medical Image Analysis and Multimodal
  Learning for Clinical Decision Support\/}, Springer, pages 282--290.
\newblock {Proc.} Fourth MICCAI International Workshop on Deep Learning in
  Medical Image Analysis (DLMIA 18).

\bibitem[{Lipton(2016)}]{lipton2016mythos}
Zachary~C Lipton. 2016.
\newblock The mythos of model interpretability.
\newblock {\em arXiv preprint arXiv:1606.03490\/} .

\bibitem[{Liu et~al.(2009)Liu, Luo, and Shah}]{liu2009recognizing}
Jingen Liu, Jiebo Luo, and Mubarak Shah. 2009.
\newblock Recognizing realistic actions from videos in the wild.
\newblock In {\em Computer vision and pattern recognition, 2009. CVPR 2009.
  IEEE conference on\/}. IEEE, pages 1996--2003.

\bibitem[{Lombrozo(2006)}]{lombrozo2006structure}
Tania Lombrozo. 2006.
\newblock The structure and function of explanations.
\newblock {\em Trends in cognitive sciences\/} 10(10):464--470.

\bibitem[{Miller(2017)}]{miller2017explanation}
Tim Miller. 2017.
\newblock Explanation in artificial intelligence: Insights from the social
  sciences.
\newblock {\em arXiv preprint arXiv:1706.07269\/} .

\bibitem[{Ribeiro et~al.(2016)Ribeiro, Singh, and Guestrin}]{ribeiro2016should}
Marco~Tulio Ribeiro, Sameer Singh, and Carlos Guestrin. 2016.
\newblock Why should i trust you?: Explaining the predictions of any
  classifier.
\newblock In {\em Proceedings of the 22nd ACM SIGKDD International Conference
  on Knowledge Discovery and Data Mining\/}. ACM, pages 1135--1144.

\bibitem[{Zhu et~al.(2007)Zhu, Mumford et~al.}]{zhu2007stochastic}
Song-Chun Zhu, David Mumford, et~al. 2007.
\newblock A stochastic grammar of images.
\newblock {\em Foundations and Trends{\textregistered} in Computer Graphics and
  Vision\/} 2(4):259--362.

\end{thebibliography}

\bibliographystyle{acl_natbib}

\end{document}